\documentclass[letterpaper,10pt,conference]{ieeeconf}

\IEEEoverridecommandlockouts
\usepackage{cite}
\usepackage{amsmath,amssymb,amsfonts}
\usepackage{graphicx}
\usepackage{textcomp}
\usepackage{xcolor}
\usepackage{booktabs}
\usepackage{multirow}
\usepackage{url}
\usepackage[hidelinks]{hyperref}
\usepackage{balance}
\usepackage{array}
\usepackage{placeins}

\title{Energy-Aware NECO for Single-Pass Pixel-wise Out-of-Distribution Detection in Semantic Segmentation}

\author{
Boyuan Zhang$^{1}$, Huanshan Huang$^{1}$, Yifei Cao$^{2,3*}$%
\thanks{$^1$Ecole Polytechnique, Institut Polytechnique de Paris, Palaiseau, France. {\tt\small firstname.lastname@polytechnique.edu}}%
\thanks{$^2$CIAD, UTBM, Université Marie et Louis Pasteur, France. {\tt\small firstname.lastname@utbm.fr}}%
\thanks{$^3$U2IS, ENSTA, Institut Polytechnique de Paris, France. {\tt\small firstname.lastname@ensta.fr}}%
\thanks{$^*$Corresponding Author. Yifei Cao was funded by the French National Research Agency (ANR) under grant agreement No ANR-23- CE10-0016.}%
}

\begin{document}

\maketitle
\thispagestyle{empty}
\pagestyle{empty}

\begin{abstract}
Reliable semantic segmentation for mobile robots requires both accurate dense prediction and robust uncertainty estimation under distribution shift. In practice, however, strong uncertainty baselines such as Monte Carlo (MC) Dropout often require repeated stochastic forward passes and are difficult to deploy on resource-constrained edge platforms. We propose Energy-Aware NEural Collapse based Out-of-distribution detection (NECO), a single-pass pixel-wise out-of-distribution (OOD) score for semantic segmentation. The method is motivated by two observations. First, preliminary experiments on the Multiple Uncertainties for Autonomous Driving (MUAD) dataset showed that uncertainty-based methods are useful for dense OOD detection, but repeated inference is expensive. Second, our earlier study on Neural Collapse (NC) and NECO suggested that purely geometric projection scores can miss simple anomalies whose features remain embedded in the dominant in-distribution subspace. We therefore combine a centered NECO-style geometric ratio computed from decoder features with a logit-based Energy score. Both components are standardized using statistics fitted on a pure in-distribution validation split and then fused by a convex combination.

We evaluate the method on the \texttt{version="small"} miniMUAD subset using true pixel-level OOD labels. The proposed hybrid score achieves an Area Under the Receiver Operating Characteristic curve (AUROC) of 0.8539, compared with 0.8280 for NECO-only, 0.8171 for Energy-only, and 0.8124 for an ensemble predictive-entropy baseline. Additional qualitative, distributional, condition-wise, and operating-point analyses show a consistent picture: the hybrid detector is stronger on the main ranking metric and across a broad operating range, while the ensemble baseline remains preferable in the extreme high-recall tail. This clarifies both the practical value and the current trade-off of the proposed single-pass design. Our code is publicly available at \url{https://github.com/boyuan-zhangx/Energy-Aware_NECO}.
\end{abstract}

\section{Introduction}
Semantic segmentation is a central dense perception task in computer vision, autonomous driving, and mobile robotics because it provides pixel-level scene understanding for downstream reasoning and control \cite{long2015fcn,ronneberger2015unet,chen2018deeplabv3plus,xie2021segformer}. Modern segmentation systems achieve strong in-distribution performance, but they are still trained under a closed-set assumption and therefore remain vulnerable to previously unseen objects, textures, scene layouts, and environmental conditions at deployment time \cite{blum2019fishyscapes,chan2021segmentmeifyoucan,bogdoll2024anovox}. This gap makes out-of-distribution detection particularly important in safety-critical applications such as autonomous driving and mobile robotics \cite{blum2019fishyscapes,chan2021entropy,rai2023unmasking}.

A common way to address this issue is to estimate predictive uncertainty. Bayesian approximations such as Monte Carlo Dropout by Gal and Ghahramani \cite{gal2016dropout} provide useful uncertainty estimates through repeated stochastic forward passes. Deep Ensembles \cite{lakshminarayanan2017ensemble} and post-hoc calibration methods such as Temperature Scaling \cite{guo2017calibration} and Local Temperature Scaling \cite{ding2021lts} further improve predictive reliability and confidence alignment. However, repeated inference or multiple independently trained models are expensive, which makes these approaches less attractive for resource-constrained edge deployment \cite{gal2016dropout,lakshminarayanan2017ensemble}.

The computational cost of repeated inference motivates single-pass post-hoc OOD scoring. Early confidence-based baselines such as Maximum Softmax Probability \cite{hendrycks2017baseline} are simple to compute. Stronger methods include ODIN by Liang \emph{et al.} \cite{liang2018odin}, Mahalanobis scoring by Lee \emph{et al.} \cite{lee2018mahalanobis}, Energy-based scoring by Liu \emph{et al.} \cite{liu2020energy}, and ViM by Wang \emph{et al.} \cite{wang2022vim}. In parallel, Neural Collapse \cite{papyan2020collapse} provides a geometric interpretation of learned representations near the terminal phase of training, and NECO by Ben Ammar \emph{et al.} \cite{ammar2024neco} explicitly exploits this structure for OOD detection through principal-component geometry. These developments suggest that feature geometry and logit evidence are both meaningful signals for OOD detection.

Dense OOD detection in semantic segmentation remains challenging because the task is spatially structured and sensitive to object boundaries, context, and scale \cite{chan2021entropy,chan2021segmentmeifyoucan,rai2023unmasking}. In addition, it is important to distinguish semantic anomalies from broader environmental covariate shifts, as the two may produce different uncertainty behaviors and influence OOD evaluation differently. Existing methods therefore span several directions, including entropy-based scoring, hybrid anomaly detection, synthetic negative data, nearest-neighbor methods, and object-centric anomaly segmentation \cite{chan2021entropy,grcic2022densehybrid,grcic2023hybridopenset,galesso2023faraway,rai2023unmasking}. This diversity indicates that no single score family is sufficient across all anomaly types.

Our design is motivated by two empirical observations. First, dense uncertainty maps are informative, but repeated inference is expensive. Second, purely geometric NECO-style scoring can miss simple anomalies that remain close to dominant in-distribution directions. We therefore propose Energy-Aware NECO, a single-pass pixel-wise OOD score for semantic segmentation. The method combines a centered NECO-style geometric ratio computed from decoder features with a logit-based Energy score. We evaluate it on miniMUAD, a lightweight subset of MUAD introduced by Franchi \emph{et al.} \cite{franchi2022muad}, and show that the hybrid score improves over both NECO-only and Energy-only under the same pixel-wise AUROC protocol.

Our contribution is threefold:
\begin{enumerate}
    \item We adapt a NECO-inspired geometric OOD score to dense decoder features for semantic segmentation.
    \item We propose a practical hybrid score that fuses centered feature-space geometry and logit-space Energy using only one forward pass.
    \item We evaluate the method on miniMUAD with true pixel-level OOD masks and show that the hybrid score improves over both individual components and a predictive-entropy ensemble baseline under the same protocol.
\end{enumerate}

\section{Related Work}

\subsection{Semantic Segmentation for Dense Perception}
Semantic segmentation has evolved from fully convolutional dense prediction to encoder--decoder and transformer-based architectures \cite{long2015fcn,ronneberger2015unet,chen2018deeplabv3plus,xie2021segformer}. These models provide strong in-distribution scene understanding, but they do not solve the open-world perception problem by themselves \cite{blum2019fishyscapes,chan2021segmentmeifyoucan}. This limitation is especially important in autonomous driving and mobile robotics, where deployment conditions often differ from the training distribution \cite{franchi2022muad,bogdoll2024anovox}.

\subsection{Uncertainty Estimation and Calibration}
Uncertainty estimation is a standard route to robust prediction under distribution shift. MC Dropout interprets dropout at inference time as approximate Bayesian inference and estimates predictive uncertainty through repeated stochastic forward passes \cite{gal2016dropout}. In computer vision, both epistemic and aleatoric uncertainty have been studied as useful signals for reliability-aware prediction \cite{kendall2017uncertainties}. Deep Ensembles and post-hoc calibration methods such as Temperature Scaling improve confidence estimation and predictive reliability without changing the main task definition \cite{lakshminarayanan2017ensemble,guo2017calibration}. For dense prediction, local calibration methods have also been proposed to better account for spatially varying confidence \cite{ding2021lts}. Despite their effectiveness, repeated inference or multiple models remain costly for edge deployment \cite{gal2016dropout,lakshminarayanan2017ensemble}.

\subsection{Dense OOD and Anomaly Segmentation}
OOD and anomaly segmentation have become established research topics for safety-critical road-scene perception \cite{blum2019fishyscapes,chan2021segmentmeifyoucan}. Fishyscapes introduced a benchmark for safe semantic segmentation in autonomous driving, while SegmentMeIfYouCan emphasized both pixel-wise and component-wise evaluation \cite{blum2019fishyscapes,chan2021segmentmeifyoucan}. On the method side, entropy-based meta classification, hybrid anomaly detection, synthetic negative data, nearest-neighbor scoring, and mask-level reasoning have all been explored for dense OOD detection \cite{chan2021entropy,grcic2022densehybrid,grcic2023hybridopenset,galesso2023faraway,rai2023unmasking}. This line of work shows that dense OOD detection often benefits from combining complementary cues rather than relying on a single scalar confidence measure \cite{grcic2022densehybrid,rai2023unmasking}.

\subsection{Single-Pass OOD Scoring and Neural Collapse}
A large family of OOD detectors operates in a post-hoc and often single-pass manner \cite{hendrycks2017baseline,liang2018odin,lee2018mahalanobis,liu2020energy,wang2022vim,ammar2024neco}. Maximum Softmax Probability is a simple confidence baseline, while ODIN improves softmax-based OOD detection through temperature scaling and input perturbation \cite{hendrycks2017baseline,liang2018odin}. Energy-based scoring later showed that logits can provide a stronger OOD signal than softmax confidence, and ViM further combined feature statistics with virtual-logit matching \cite{liu2020energy,wang2022vim}. Mahalanobis-based methods also demonstrated that class-conditional feature geometry can be useful for post-hoc OOD detection \cite{lee2018mahalanobis}. On the representation side, Neural Collapse describes a family of geometric regularities emerging near the terminal phase of training, and NECO uses this geometry to construct a post-hoc OOD score \cite{papyan2020collapse,ammar2024neco}. Our method follows this general direction, but adapts the idea to dense decoder features and explicitly complements the geometric score with Energy.

\section{Background and Motivation}
The motivation above is developed here in more detail. First, on MUAD, uncertainty-based dense prediction is meaningful. In our preliminary exploration, Deep Ensemble improved Expected Calibration Error (ECE) relative to a single model, and predictive-entropy maps from MC Dropout produced sensible spatial patterns, with uncertainty concentrated around semantic boundaries for in-distribution scenes and more globally elevated on OOD scenes. This observation is consistent with the general use of uncertainty estimation for reliability-aware prediction in computer vision and dense perception \cite{gal2016dropout,kendall2017uncertainties,lakshminarayanan2017ensemble,ding2021lts}. At the same time, it also highlights the practical drawback of repeated inference for edge deployment \cite{gal2016dropout,lakshminarayanan2017ensemble}.

Second, in our earlier OOD and Neural Collapse exploration on image classification, purely geometric NECO-style scores worked better on high-energy, directionally distinct anomalies than on simple anomalies that remained close to the learned in-distribution subspace. This observation is consistent with the broader intuition that different OOD types may require different scoring cues and that hybrid feature-logit methods can be more robust than purely geometric or purely confidence-based methods alone \cite{wang2022vim,ammar2024neco,grcic2022densehybrid,grcic2023hybridopenset}. It therefore suggests that geometry alone is not always sufficient and that a complementary score based on logit magnitude or evidence may recover part of the missing signal \cite{liu2020energy,wang2022vim}.

\section{Energy-Aware NECO Formulation}

\subsection{Problem Setting}
An input image is denoted by $x \in \mathbb{R}^{H \times W \times 3}$, where $H$ and $W$ are the image height and width. The segmentation network produces pixel-wise logits $z(x) \in \mathbb{R}^{H \times W \times C}$ and decoder features $h(x) \in \mathbb{R}^{H \times W \times d}$, where $C$ is the number of in-distribution (ID) semantic classes and $d$ is the decoder feature dimension. Our goal is to assign to each valid pixel an OOD score such that larger values indicate stronger OOD evidence.

\subsection{Dense NECO Adaptation}
The original NECO score measures how strongly a feature projects onto the principal ID subspace \cite{ammar2024neco}. In our dense adaptation, we collect decoder pixel features from a pure ID split and compute their empirical mean $\mu$ together with a principal-component basis $P$. For a pixel feature $h(x)$, the centered geometric ratio is defined as
\begin{equation}
    \mathrm{NECO}(x)=\frac{\|P(h(x)-\mu)\|_2}{\|h(x)-\mu\|_2+\epsilon},
    \label{eq:neco}
\end{equation}
where $\epsilon>0$ is a numerical-stability constant.

This formulation is motivated by the observation that deep representations tend to form structured class-wise geometries near the end of training, consistent with the Neural Collapse phenomenon \cite{papyan2020collapse}.This formulation is not an exact reproduction of the original NECO score. The modification is intentional: the original formulation was designed for classification representations, whereas our setting uses dense pixel-level decoder features. The centering step and projection pipeline therefore provide a stable and tractable adaptation to dense segmentation features while preserving the original geometric intuition.

\subsection{Energy Term}
For the same pixel, we compute the logit-based Energy score \cite{liu2020energy}
\begin{equation}
    \mathrm{Energy}(x)=T \cdot \log \sum_{c=1}^{C} \exp\left(\frac{\mathrm{logit}_c(x)}{T}\right),
    \label{eq:energy}
\end{equation}
where $c$ indexes the semantic classes, $\mathrm{logit}_c(x)$ denotes the logit of class $c$ at the considered pixel, and $T>0$ is the temperature parameter. The Energy term captures the magnitude and evidence structure of the logits. It complements the geometric ratio because some anomalous pixels may remain aligned with dominant ID directions while still producing atypical logit evidence.

\subsection{Validation-Fitted Standardization}
The raw NECO and Energy values operate on different scales. We therefore standardize them using a pure ID validation split. Let $\mu_{\mathrm{NECO}}$, $\sigma_{\mathrm{NECO}}$, $\mu_{\mathrm{Energy}}$, and $\sigma_{\mathrm{Energy}}$ denote the empirical mean and standard deviation of the two raw scores on that split. We define
\begin{equation}
    Z_{\mathrm{NECO}}=\frac{\mathrm{NECO}(x)-\mu_{\mathrm{NECO}}}{\sigma_{\mathrm{NECO}}}
\end{equation}
and
\begin{equation}
    Z_{\mathrm{Energy}}=\frac{\mathrm{Energy}(x)-\mu_{\mathrm{Energy}}}{\sigma_{\mathrm{Energy}}}.
\end{equation}
Only ID data are used for score fitting, which matches a realistic OOD protocol in which unknown categories are unavailable at fitting time \cite{hendrycks2017baseline,liu2020energy}.

\subsection{Hybrid OOD Score}
The fused ID-oriented score is defined as
\begin{equation}
    S_{\mathrm{ID}}(x)=\alpha \cdot Z_{\mathrm{NECO}} + (1-\alpha)\cdot Z_{\mathrm{Energy}},
    \label{eq:sid}
\end{equation}
where $\alpha \in [0,1]$ controls the trade-off between the geometric and logit-based terms. In our implementation, $\alpha=0.6$, which gives slightly more weight to the geometric component, as it empirically provides a more stable reference under dense spatial representations.
\begin{equation}
    S_{\mathrm{OOD}}(x)=-S_{\mathrm{ID}}(x),
    \label{eq:sood}
\end{equation}
so that larger values correspond to stronger OOD evidence.

\subsection{Complementarity of Geometry and Energy}
The proposed fusion is motivated by complementary failure modes in post-hoc OOD detection. If an anomalous pixel deviates directionally from the learned ID principal subspace, the geometric term can detect it, which matches the Neural Collapse intuition behind NECO \cite{papyan2020collapse,ammar2024neco}. If an anomalous pixel remains close to the dominant ID subspace but produces atypical logit magnitude or evidence, the Energy term can still provide useful discrimination \cite{liu2020energy}. By combining both signals, the proposed hybrid score can handle a wider range of dense anomalies under a single forward pass.

\section{Experimental Setup}

\subsection{Dataset}
We evaluate on MUAD \cite{franchi2022muad}. In the present implementation, we use the \texttt{version="small"} miniMUAD subset. This controlled subset is intentionally used to focus the evaluation on semantic anomaly detection while reducing the influence of severe environmental covariate shifts such as extreme lighting and weather conditions. We therefore focus our analysis on the scenes and proxy conditions that are available within this setup.

In our protocol, the OOD mask is a true pixel-level label mask, not a proxy. The segmentation model is trained with $C=19$ ID classes. Pixels with labels greater than or equal to the class count are treated as OOD labels in the evaluation split, while ignored pixels are removed from evaluation.

\subsection{Preprocessing and Score Fitting Split}
All images and masks are resized to $(256,512)$. During training, we apply random horizontal flipping, convert images to \texttt{float32}, and normalize channels with ImageNet statistics. Labels with index $\geq 19$ are remapped to the ignore value $255$, so rare or anomalous classes do not contribute to the segmentation loss.

All score statistics are fitted on a pure ID validation split. Concretely, the validation loader is used to estimate the PCA basis $P$, the feature mean $\mu$, and the mean and standard deviation of the raw NECO and Energy scores. No OOD labels or OOD score statistics are used during score fitting.

\subsection{Baselines}
We compare our method with the following baselines:
\begin{itemize}
    \item Ensemble predictive entropy,
    \item NECO-only,
    \item Energy-only,
    \item Hybrid (NECO + Energy).
\end{itemize}
The ensemble baseline uses predictive entropy as the OOD score. It is therefore a reasonable uncertainty reference, but it is not the same scoring family as the proposed hybrid.

\subsection{Metrics}
We report pixel-wise AUROC as the primary quantitative metric, which is standard in OOD and anomaly segmentation benchmarks \cite{blum2019fishyscapes,chan2021segmentmeifyoucan}. We also report the mean score on ID pixels and the mean score on OOD pixels to illustrate score separation. In addition, we analyze the false positive rate at 95\% true positive rate (FPR95), the false positive rate at 98\% true positive rate (FPR98), and the shape of the ROC curve, because operating-point behavior matters for safety-oriented dense OOD detection.

\section{Results}
Table~\ref{tab:main} summarizes the main quantitative comparison.

\begin{table}[t]
    \caption{Comparison of OOD scores on miniMUAD under the same pixel-wise AUROC protocol. Higher AUROC is better.}
    \label{tab:main}
    \centering
    \begin{tabular}{lccc}
        \toprule
        Method & AUROC & ID mean & OOD mean \\
        \midrule
        Ensemble (predictive entropy) & 0.8124 & 0.3021 & 0.8271 \\
        NECO-only & 0.8280 & 0.3916 & 2.1789 \\
        Energy-only & 0.8171 & -0.0601 & 1.2303 \\
        Hybrid (NECO + Energy) & \textbf{0.8539} & 0.2110 & 1.7995 \\
        \bottomrule
    \end{tabular}
\end{table}

The hybrid score achieves the best AUROC, improving over both NECO-only and Energy-only. This supports the intended complementarity between feature-space geometry and logit-space evidence. The separation between ID and OOD mean scores is also clear, which suggests cleaner thresholding than with either individual component.

The comparison with the ensemble baseline should be interpreted as a comparison of deployment-oriented detection effectiveness rather than as a claim that both methods quantify the same notion of uncertainty. The ensemble baseline is a multi-model predictive-uncertainty estimate, whereas the proposed method is a single-pass fused OOD score. The main result is therefore not that the two methods are theoretically equivalent, but that the proposed single-pass detector remains competitive with a stronger but more expensive uncertainty baseline.

\section{Qualitative and Distributional Analysis}
The main quantitative result in Table~\ref{tab:main} is complemented by four figures that explain why the hybrid detector works, where the gain comes from, and why a geometric term alone is not sufficient.

\subsection{Calibration Evidence from the Baseline Segmenter}
\begin{figure}[t]
    \centering
    \includegraphics[width=\linewidth]{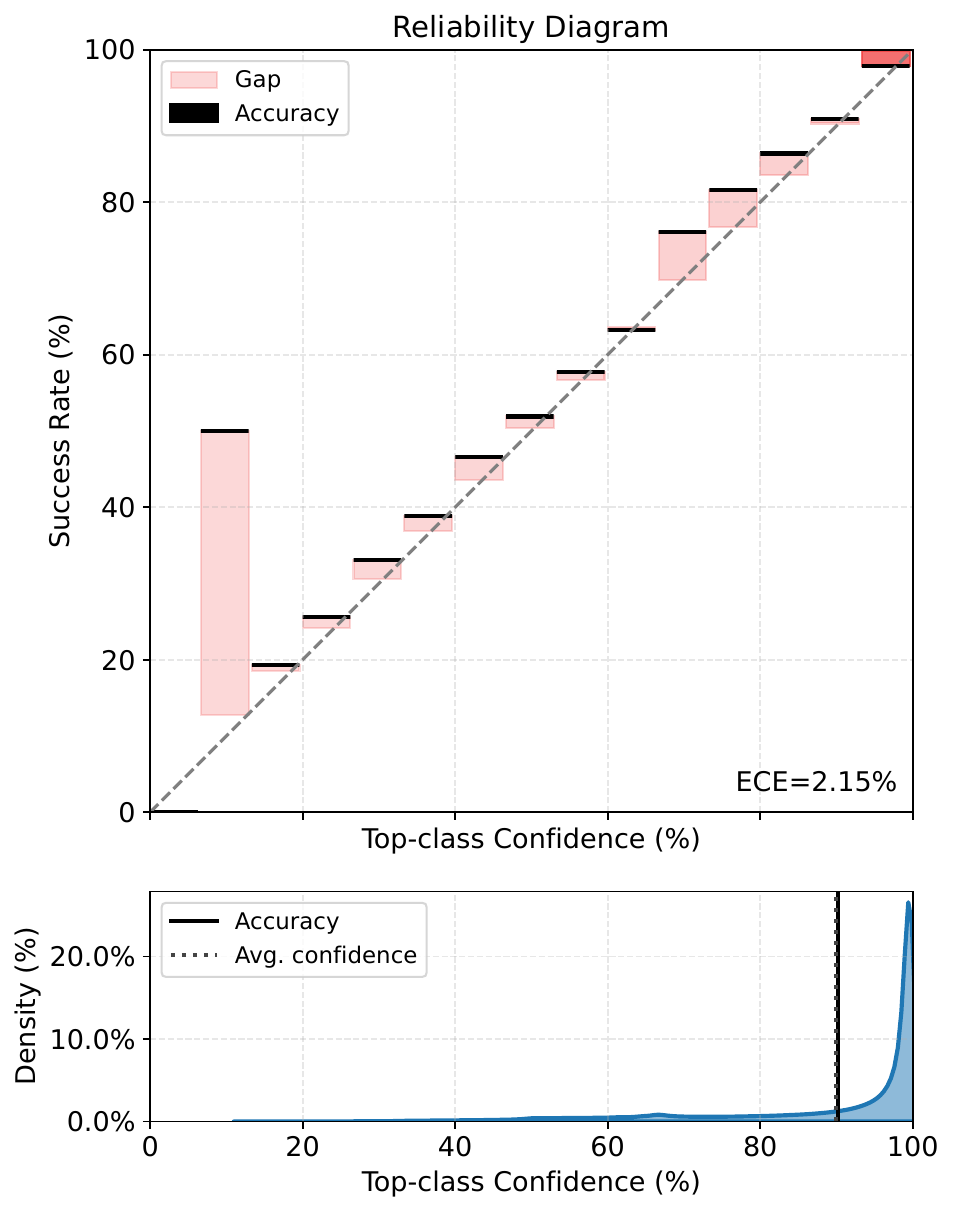}
    \caption{Calibration evidence for the baseline segmentation model. The reliability diagram and confidence histogram show that the backbone already exhibits non-trivial confidence structure, which motivates uncertainty-aware dense OOD analysis on top of this predictor.}
    \label{fig:ece_old}
\end{figure}

Figure~\ref{fig:ece_old} shows the calibration behavior of the segmentation backbone. The reliability diagram indicates that the baseline predictor is not arbitrarily overconfident, and the confidence histogram remains concentrated near the high-confidence region with a limited calibration gap. This figure establishes that uncertainty-like outputs from the backbone already contain meaningful structure, which makes it reasonable to compare entropy-based and post-hoc OOD scores on top of the same predictor.

\subsection{Multi-Scene Qualitative OOD Maps}
\begin{figure*}[t]
    \centering
    \includegraphics[width=0.8\textwidth]{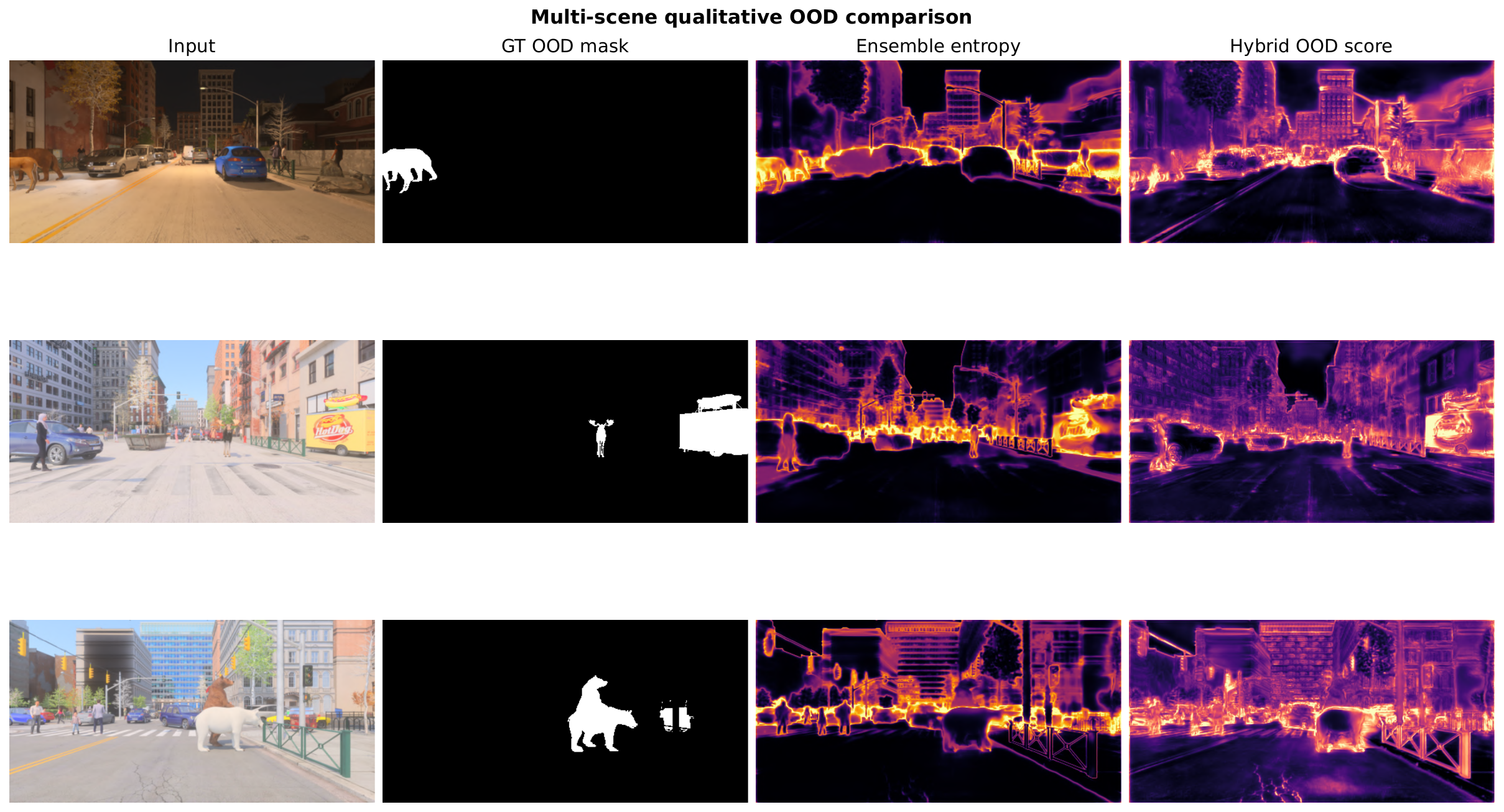}
    \caption{Multi-scene qualitative OOD comparison on three representative miniMUAD scenes. Each row shows the input image, the ground-truth OOD mask, the ensemble predictive-entropy map, and the proposed hybrid OOD score. Across the three scenes, the hybrid detector produces more spatially continuous responses over the annotated OOD objects, while the ensemble baseline remains more edge-dominated.}
    \label{fig:qualitative_h1}
\end{figure*}

Figure~\ref{fig:qualitative_h1} replaces the previous single-scene qualitative comparison with three representative miniMUAD scenes. Across these scenes, the hybrid detector produces more spatially continuous responses over the annotated OOD objects, with fewer internal holes and less fragmentation than the ensemble entropy maps. The ensemble baseline remains strong around generic scene edges, but its high-score regions are more diffuse and more tightly coupled to boundary structure. The figure therefore supports a more precise statement than the original draft: the advantage of Hybrid is not only numerical, but also structural at the object level.

\begin{figure*}[!t]
    \centering
    \includegraphics[width=\linewidth]{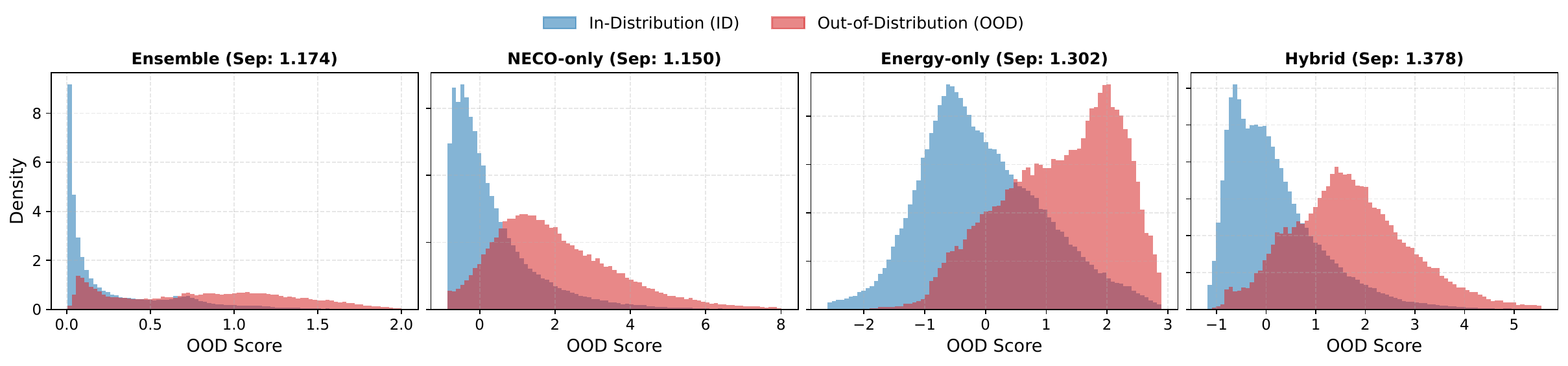}
    \caption{Distribution of ID and OOD scores for the compared methods. The hybrid score shows the clearest reduction of ID/OOD overlap, which is consistent with its AUROC advantage in Table~\ref{tab:main}.}
    \label{fig:distribution_new}
\end{figure*}

\subsection{Score Distribution Analysis}
Figure~\ref{fig:distribution_new} explains the AUROC improvement in Table~\ref{tab:main}. The hybrid detector shows the clearest reduction of overlap between ID and OOD score distributions. Relative to the ensemble baseline, the OOD mass is shifted farther toward larger scores while the ID mass remains more concentrated at lower values. This is exactly the distributional pattern that supports better global ranking and more stable threshold selection.

\subsection{Feature-Geometry Failure Case}
\begin{figure*}[!t]
    \centering
    \includegraphics[width=0.8\textwidth]{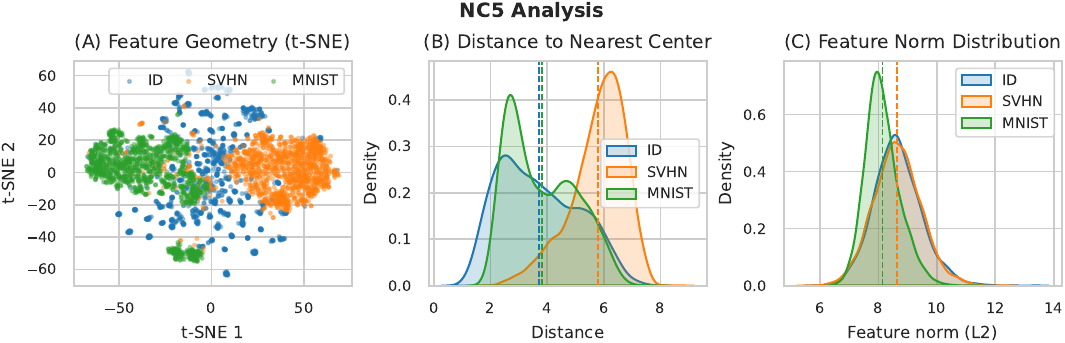}
    \caption{Feature-geometry failure case from the earlier OOD and Neural Collapse exploration. The figure illustrates why a purely geometric detector may miss simple anomalies that remain embedded in dominant ID directions, which motivates the addition of the Energy term.}
    \label{fig:nc5_old}
\end{figure*}

Figure~\ref{fig:nc5_old} clarifies why the Energy term is necessary. It illustrates a representative failure case of a purely geometric detector: some simple anomalies remain embedded in dominant in-distribution directions and are therefore not cleanly separated by subspace projection alone. The figure justifies the design decision in Section~IV: if geometry alone can miss a simple anomaly mode, then a complementary logit-based cue becomes well motivated.

\section{Discussion and Future Work}
The main conclusion from Table~\ref{tab:main} is not merely that the hybrid score attains the largest AUROC. The more important point is that the hybrid score improves over both NECO-only and Energy-only, which supports the intended complementarity between feature-space geometry and logit-space evidence. This interpretation is consistent with prior work showing that different OOD detectors capture different aspects of distribution shift and that hybrid dense OOD methods can benefit from combining multiple cues \cite{liu2020energy,wang2022vim,grcic2022densehybrid,grcic2023hybridopenset}.

This observation is also consistent with the design logic of our method. The centered NECO term captures subspace alignment with the learned ID representation. The Energy term captures abnormal activation magnitude and logit evidence. Some anomalies are well described by geometry, while others are better described by Energy. The hybrid combines both under a single-pass constraint.

The result is practically meaningful for edge deployment. Uncertainty-based baselines such as predictive entropy from ensembles or MC Dropout are useful, but they are more expensive at inference time \cite{gal2016dropout,lakshminarayanan2017ensemble}. By contrast, the proposed score reuses features and logits that are already available in a standard forward pass.

\subsection{Operating-Point Trade-Off in the High-TPR Tail}
\begin{figure}[!tb]
    \centering
    \includegraphics[width=0.8\linewidth]{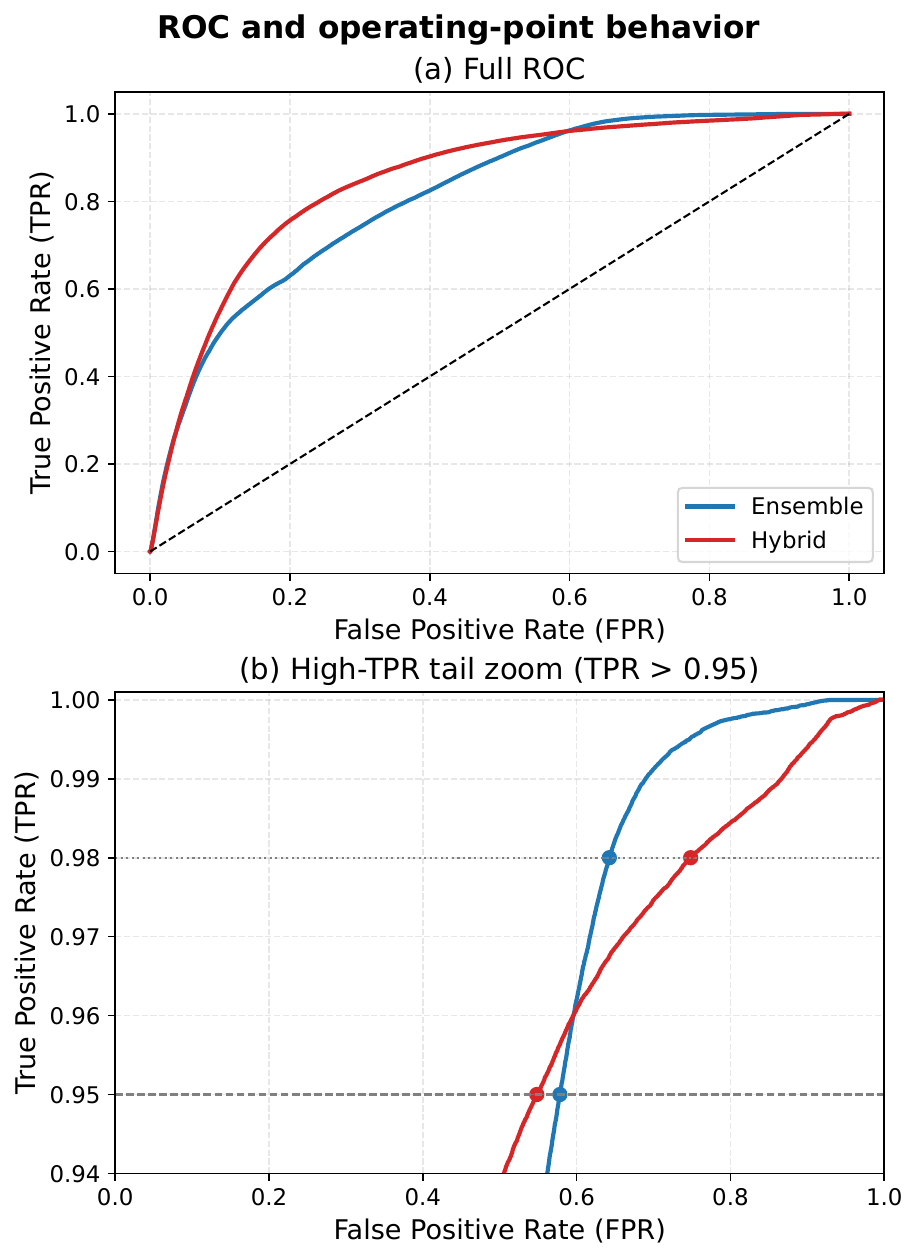}
    \caption{ROC and operating-point behavior. Top: full ROC curve. Bottom: zoom on the high-TPR tail. Hybrid dominates over a broad operating range, but the ensemble baseline becomes preferable near the extreme recall regime.}
    \label{fig:h3_roc}
\end{figure}

\subsection{Condition-Wise Robustness Under Proxy Conditions}
\begin{figure*}[!t]
    \centering
    \includegraphics[width=0.88\textwidth]{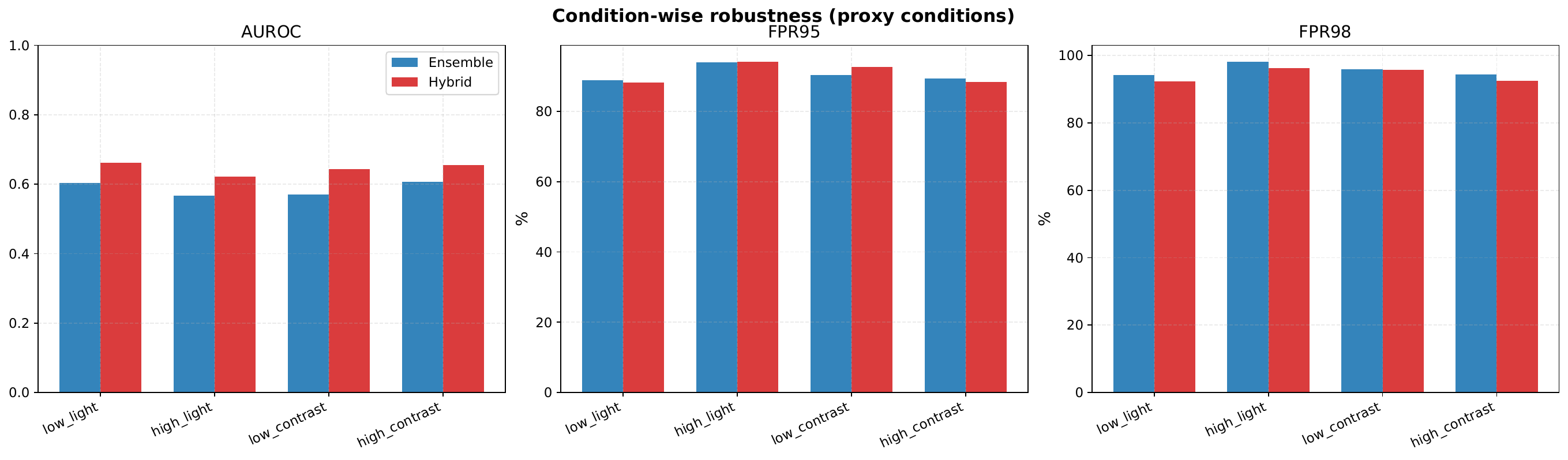}
    \caption{Condition-wise robustness under four proxy conditions available in miniMUAD. Hybrid improves AUROC across all four subsets. FPR95 remains broadly comparable. The condition-wise FPR98 bars stay close as well, suggesting that the most severe weakness of Hybrid is not a uniform per-condition degradation, but a high-recall tail effect revealed by the global ROC analysis.}
    \label{fig:h2_conditionwise}
\end{figure*}

Figure~\ref{fig:h3_roc} makes the trade-off explicit. In the full ROC curve, the hybrid detector dominates the ensemble baseline over a broad range of false-positive rates, which is consistent with the AUROC result in Table~\ref{tab:main}. However, the tail zoom reveals a crossover in the high-TPR region. Around the very high recall regime, the ensemble baseline becomes preferable: at the marked $\mathrm{TPR}=0.98$ operating point, the ensemble reaches the target recall with a visibly lower false-positive rate than the hybrid detector. In contrast, at $\mathrm{TPR}=0.95$, the two methods are closer and Hybrid remains competitive.

This figure clarifies the main trade-off of the paper. The proposed hybrid score is stronger on the global ranking metric and on much of the operating range, but it is less well controlled in the extreme high-recall tail. A plausible explanation is that the Energy component produces a heavier high-score tail on hard ID pixels. When the threshold is lowered to capture the final few percent of OOD pixels, these difficult ID pixels cross the decision boundary in large numbers. By contrast, the ensemble baseline benefits from variance reduction through multi-model averaging, which yields better false-positive control in the most conservative regime.

Figure~\ref{fig:h2_conditionwise} analyzes four proxy conditions available in miniMUAD: \texttt{low\_light}, \texttt{high\_light}, \texttt{low\_contrast}, and \texttt{high\_contrast}. Three observations matter. First, Hybrid remains above Ensemble in AUROC for all four subsets. Second, FPR95 remains broadly comparable, with only modest condition-dependent differences. Third, the condition-wise FPR98 bars remain close as well. This means that the operating-point weakness of the hybrid detector does not appear as a uniform collapse across all proxy conditions. Instead, the more severe weakness emerges in the extreme high-recall tail of the global ROC curve.

The current study also points to several natural future directions. First, a local spatial smoothing stage or lightweight structured post-processing, such as average pooling, Gaussian smoothing, or a compact Conditional Random Field (CRF), could suppress isolated false positives on hard ID pixels. Second, outlier exposure or virtual outlier synthesis could explicitly tighten the ID energy tail during training. Third, logit normalization or a distance-based alternative to raw Energy, such as Mahalanobis or nearest-neighbor scoring in decoder space, could improve behavior near extreme operating points. Fourth, a head-only lightweight ensemble, for example dropout restricted to the prediction head, could recover part of the variance reduction of deep ensembles without paying their full computational cost.

\section{Conclusion}
We presented Energy-Aware NECO, a single-pass pixel-wise OOD score for semantic segmentation. The method fuses a centered NECO-style geometric ratio with a logit-based Energy score, using score statistics fitted on a pure ID validation split. On miniMUAD with true pixel-level OOD masks, the proposed hybrid achieves the best AUROC among the compared methods.

The broader conclusion is nuanced. The hybrid detector offers a strong efficiency--performance compromise for dense OOD detection, with cleaner global ranking, stronger qualitative object-level responses, and robust behavior across several proxy-condition splits. At the same time, the ROC-tail analysis shows that repeated-inference uncertainty estimation still retains an advantage in the most conservative high-recall regime. This makes the proposed method a promising deployment-oriented detector rather than a universal replacement for ensemble-based uncertainty estimation.

\balance
\FloatBarrier 

\end{document}